\documentclass[runningheads]{llncs}
\usepackage{graphicx}
\usepackage{tikz}
\usepackage{comment}
\usepackage{amsmath,amssymb} % define this before the line numbering.
\usepackage{color}

\usepackage[accsupp]{axessibility}  % Improves PDF readability for those with disabilities.

\usepackage{booktabs}
\usepackage{csquotes}
\usepackage{tabularx}
\usepackage{epsfig}
\usepackage{subfiles}
\usepackage{float,xspace}
\usepackage{subcaption}
\newcolumntype{Y}{>{\centering\arraybackslash}X}
\newcolumntype{R}{>{\raggedright\arraybackslash}X}

\usepackage{comment}
\usepackage{algorithm,algorithmic,color}
\usepackage{boldline} 
\usepackage{dsfont}
\usepackage{color,soul}
\usepackage{balance}
\usepackage{multirow}
\usepackage{multicol}

\usepackage{diagbox}
\usepackage{xcolor}
\usepackage{enumitem}

\makeatletter
\DeclareRobustCommand\onedot{\futurelet\@let@token\@onedot}
\def\@onedot{\ifx\@let@token.\else.\null\fi\xspace}

\def\eg{\emph{e.g}\onedot} 
\def\ie{\emph{i.e}\onedot}

\def\etal{\emph{et al}\onedot}
\makeatother

\newcommand{\bR}{\mathbb{R}\xspace}
\newcommand{\cL}{\mathcal{L}\xspace}
\newcommand{\cM}{\mathcal{M}\xspace}
\newcommand{\compressor}[0]{FeatComp\xspace}
\newcommand{\supervision}[0]{supervision size\xspace}

\newcommand{\taskind}[0]{R-only\xspace}
\newcommand{\plusrecons}[0]{FeatComp+R\xspace}

\newif\ifshowcomments
\newcommand{\done}[1]{}
\newcommand{\shiyuandone}[1]{}
\newcommand{\gunnardone}[1]{}
\newcommand{\robinsondone}[1]{}
\newcommand{\shihfudone}[1]{}

\showcommentstrue  % comment this out to hide comments
\ifshowcomments
    \newcommand{\shiyuan}[1]{{\color{orange}[shiyuan: #1]}}
    \newcommand{\gunnar}[1]{{\color{violet}[gunnar: #1]}}
    \newcommand{\robinson}[1]{{\color{teal}[robinson: #1]}}
    \newcommand{\shihfu}[1]{{\color{green}[shih-fu: #1]}}
\else
    \newcommand{\shiyuan}[1]{}
    \newcommand{\gunnar}[1]{}
    \newcommand{\robinson}[1]{}
    \newcommand{\shihfu}[1]{}
\fi

\begin{document}
% \linenumbers
\pagestyle{headings}
\mainmatter
\def\ECCVSubNumber{****}  % Insert your submission number here

%\title{Video in 10 bits: Task-Specific Feature Compression} % Replace with your title
\title{Video in 10 Bits: \\Few-Bit VideoQA for Efficiency and Privacy}

\begin{comment}
\titlerunning{ECCV-22 submission ID \ECCVSubNumber} 
\authorrunning{ECCV-22 submission ID \ECCVSubNumber} 
\author{Anonymous ECCV submission}
\institute{Paper ID \ECCVSubNumber}
\end{comment}
%******************

%\begin{comment}
\titlerunning{Few-Bit VideoQA}
\author{Shiyuan Huang\inst{1}%\orcidID{0000-1111-2222-3333} 
\and
Robinson Piramuthu\inst{2}%\orcidID{1111-2222-3333-4444} 
\and
Shih-Fu Chang\inst{1}
%\orcidID{2222--3333-4444-5555}
\and\\
Gunnar A. Sigurdsson\inst{2}
}
\authorrunning{S. Huang et al.}
\institute{Columbia University \and
Amazon Alexa AI
%\email{\{shiyuan.h,sc250\}@columbia.edu}\\
%\email{shiyuan.h@columbia.edu}
%\url{http://www.springer.com/gp/computer-science/lncs} 
%\email{\{gsig,robinpir\}@amazon.com}
}
%\end{comment}
%******************
\maketitle

\begin{abstract}
In Video Question Answering (VideoQA), answering general questions about a video requires its visual information. Yet, video often contains redundant information irrelevant to the VideoQA task. For example, if the task is only to answer questions similar to ``Is someone laughing in the video?'', then all other information can be discarded.
This paper investigates how many bits are really needed from the video in order to do VideoQA by introducing a novel \emph{Few-Bit VideoQA} problem, where the goal is to accomplish VideoQA with few bits of video information (e.g., 10 bits). We propose a simple yet effective task-specific feature compression approach to solve this problem. Specifically, we insert a lightweight  \textbf{Feat}ure \textbf{Comp}ression Module (\textbf{\compressor}) into a VideoQA model which learns to extract task-specific tiny features as little as 10 bits, which are optimal for answering certain types of questions. 
We demonstrate more than 100,000-fold storage efficiency over MPEG4-encoded videos and 1,000-fold over regular floating point features, with just $2.0{-}6.6\%$ absolute loss in accuracy, which is a surprising and novel finding.
Finally, we analyze what the learned tiny features capture and demonstrate that they have eliminated most of the non-task-specific information, and introduce a Bit Activation Map to visualize what information is being stored.
This decreases the privacy risk of data by providing k-anonymity and robustness to feature-inversion techniques, which can influence the machine learning community, allowing us to store data with privacy guarantees while still performing the task effectively.

\end{abstract}

%%%%%%%%% BODY TEXT
\section{Introduction}
\label{sec:intro}

\begin{figure}[t]
\centering
\includegraphics[width=0.8\linewidth]{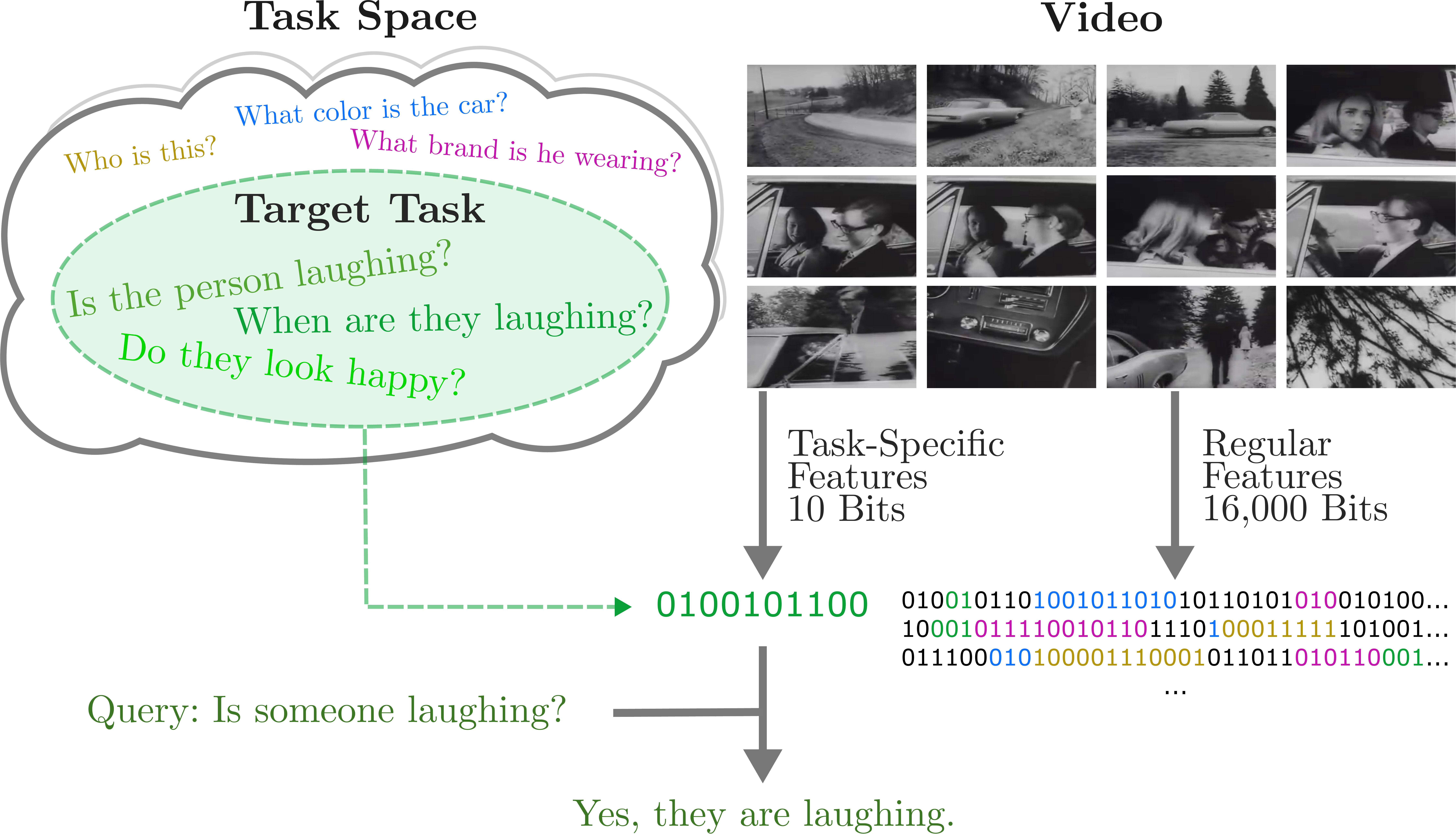}
\vspace{-2mm}
\caption{
We introduce the problem of \emph{Few-Bit VideoQA} where only few bits of video information is allowed to do VideoQA tasks. 
Our proposed method compresses the features inside a neural network to an extreme degree: A video input of 1 MB
can be compressed down to 10 bits, and still solve common question-answering tasks with a high degree of accuracy, such as ``Is someone laughing?''. This has both storage and privacy implications.
Here the \emph{Target Task} is all questions related to \emph{laughing}, a subset of all possible tasks in the \emph{Task Space}. \emph{Regular Features} have additional task-irrelevant information visualized with questions and bits of corresponding color.
\emph{Regular Features} corresponds to 512 floating point features from a state-of-the-art video network, such as~\cite{tran2018closer}.
}
% TODO explain regular features, or floating point features, maybe say where they are from 
\label{fig:concept}
\end{figure}

Video data exemplifies various challenges with machine learning data: It is large and has privacy issues (\eg faces and license plates). For example, an autonomous robot driving around may collect gigabytes of video data every hour, quickly filling up all available storage with potentially privacy-sensitive information. Yet, video is useful for a variety of computer vision applications, \eg action detection~\cite{lin2019bmn}, object tracking~\cite{minderer2019unsupervised}, and video question answering (VideoQA)~\cite{xu2017video,Le_2020_CVPR}. With increasing adoption of computer vision applications, and mobile devices, efficiently storing video data becomes an increasingly important problem.

%* Many problems don’t need pixel level information. In fact, some cases require only few bits. The video side takes a lot of space and has privacy concerns. Same for features.
VideoQA~\cite{xu2017video,Le_2020_CVPR} is a general machine learning task that requires analyzing video content to answer a general question. As such, it contains elements of multiple video problems, such as classification, retrieval, and localization. The questions in VideoQA are not all available beforehand, which means that the model needs to extract general semantic information that is still useful enough to answer a diverse set of questions, such as any question about the task ``Human Activities'', for example.
Concretely, the system may have been trained on queries including ``Is the person laughing?'', and at inference time, the system needs to record and store a video, and may later receive the query ``Do they look happy?'' which was not seen at training time. In contrast with classification problems where the system could potentially store the answer to all possible queries, a VideoQA system needs to keep enough information to answer any query about the video.

While early video analysis used hand-crafted visual descriptors to represent videos, recent advances in deep neural networks are able to learn highly abstracted features~\cite{carreira2017quo,miech19howto100m,ILSVRC15}. These features still contain more information than what is actually needed in a specific task. For example, a popular video encoder network ResNet3D~\cite{tran2018closer} extracts a 512-dim floating number feature, which requires 16,384 bits. If the task is to only answer questions about the happiness of a person, such as ``do people look happy'', we theoretically only need 1-bit of information to accomplish that task by encoding the existence of person + laughing. A single 16,384-bit feature encodes much more information than required to answer this question. Moreover, it is hard to interpret what information is captured by continuous features, and hence hard to guarantee that the features do not contain any sensitive information, which may carry privacy concerns.

%* Most recent work improves by learning stronger vision features or designing better multi-modal interaction. We instead want to investigate this problem from another direction - We investigate how many bits are really needed from the video in current VL tasks
Much of recent work on VideoQA focuses on learning stronger vision features, improved architectures, or designing better multi-modal interaction~\cite{feichtenhofer2019slowfast,sun2019videobert,li2020hero,miech2020end,lei2021less}. This paper instead investigates how many bits are really needed from video in current VideoQA tasks. 
To this end, we introduce a novel ``\emph{Few-Bit VideoQA}'' problem, which aims to accomplish VideoQA where only few bits of information from video are allowed. To our knowledge, the study of few-bit features is an understudied problem, with applications to storing and cataloging large amounts of data for use by machine learning applications.

We provide a simple yet effective task-specific compression approach towards this problem. Our method is inspired by recent learning-based image and video coding~\cite{lu2019dvc,liu2020mlvc,hu2020imporving,abdelaziz2019neural,feng2020learned,Hu_2021_CVPR,Chadha_2021_CVPR,toderici2015variable,toderici2017full,wu2018video}, which learns low-level compression with the goal of optimizing visual quality. In contrast, our method looks at compressing high-level features, as shown in Figure~\ref{fig:concept}.
Given a video understanding task, we compress the deep video features into few bits (e.g., 10 bits) to accomplish the task. Specifically, we utilize a generic \textbf{Feat}ure \textbf{Comp}ression module (\textbf{\compressor}), which can be inserted into neural networks for end-to-end training. \compressor learns compressed binarized features that are optimized towards the target task. 
In this way, our task-specific feature compression can achieve a high compression ratio and also address the issue of privacy. This approach can store large amounts of features on-device or in the cloud, limiting privacy issues for stored features, or transmitting only privacy-robust features from a device.
Note that this work is orthogonal to improvements made by improved architectures, shows insights into analyzing how much video data is needed for a given VideoQA dataset, and provides a novel way to significantly optimize storage and privacy risks in machine learning applications.

\begin{table}[t]
\caption{Overview of different levels of feature storage and privacy limitations.
%\caption{Different levels of feature storage and their privacy limitations.
\done{Are we replacing Original Data with MPEG?}
}
%\begin{tabularx}{\textwidth}{Xlll}
\resizebox{\textwidth}{!}{%
%\begin{tabularx}{\linewidth}{llll}
\begin{tabular}{llll}
\toprule
 & Data Amount & Can Identify User? & Can Discriminate User?  \\ \midrule
% Original Data & 630KB (e.g. MPEG4 video) & Yes & Yes \\
Original Data & 1,000,000 bits (e.g. MPEG4 video) & Yes & Yes \\
Regular Feature & 16,000 bits (e.g. 512 floats) & Yes (Can be inverted) & Yes  \\
Regular Compressed Feature & 100 bits & No & Yes  \\
Task-Compressed Feature & 10 bits & No & No (k-Anonymity~\cite{Samarati98protectingprivacy})  \\ \bottomrule
%\end{tabularx}
\end{tabular}
}
\label{tbl:levels}
\end{table}

%* Why 10 bits?
10-bits is a concrete number we use throughout this paper to demonstrate the advantage of tiny features. 
While 10-bits seems like too little to do anything useful, we surprisingly find that predicting these bits can narrow down the solution space enough such that the model can correctly pick among different answers in VideoQA~\cite{xu2017video,Le_2020_CVPR}. Different tasks require different number of bits, and we see in Section~\ref{sec:experiments} that there are different losses of accuracy for different tasks at the same number of bits.
Storing only 10 bits, we can assure that the stored data does not contain any classes of sensitive information that would require more than 10 bits to be stored. For example, we can use the threshold of 33 bits (8 billion unique values) as a rule-of-thumb threshold where the features stop being able to discriminate between people in the world. In Table~\ref{tbl:levels} we show different levels of features and how they compare to this threshold. At the highest level, \emph{Original Data}, we would have the full image or video with all their privacy limitations.\footnote{In the MSRVTT-QA dataset the videos are 630KB on average, which consists of $320{\times} 240$-resolution videos, $15$s on average.} Even with feature extraction, various techniques exist to invert the model and reconstruct the original data~\cite{fredrikson2015model,mahendran2015understanding}. Using \emph{Regular Compressed Features} (\ie off-the-shelf compression on features in Section~\ref{sec:tiny_data} or task-independent compression in Section~\ref{sec:videoqa_exp}) would still pose some privacy threats. Our \emph{Task-Compressed Features}, provide privacy guarantees from their minimal size.

%* We systematically investigate the problem and analyze its application-wise benefits (privacy, storage, etc).
We experiment on public VideoQA datasets to analyze how many bits are needed for VideoQA tasks. 
%We also provide a study of what information the compression learned, and apply it to various applications to validate its effectiveness. 
In our experiments, the ``\textbf{Task}'' is typically question types in a specific dataset, such as TGIF-Action~\cite{jang2017tgif}, but in a practical system this could be a small set of the most common queries, or even a single type of question such as ``do people look happy in the video?''. Note that while the ``Task'' of all questions in a single dataset might seem very diverse and difficult to compress, the task is much more narrow than any possible question about any information in the video, such as ``What color is the car?'', ``What actor is in the video?'', etc. %or ``What brand is he wearing?''.
In summary, our contributions can be summarized as:
\begin{enumerate}[itemsep=2pt,topsep=0pt,parsep=0pt]
    \item We introduce a novel Few-Bit VideoQA problem, where only few bits of video information is used for VideoQA; and we propose a simple yet effective task-specific feature compression approach that learns to extract task-specific tiny features of very few bits.
    
    \item Extensive study of how many bits of information are needed for VideoQA. We demonstrate that we lose just $2.0\% {-} 6.6\%$ in accuracy using only 10 bits of data. It provides a new perspective of understanding how much visual information helps in VideoQA.
    
    \item We validate and analyze the task specificity of the learned tiny features,  and  demonstrate their storage efficiency and privacy advantages. 

\end{enumerate}

The outline of the paper is as follows. In Section~\ref{sec:related}, we review related work in VideoQA, image/video coding and deep video representations. In Section~\ref{sec:method} we introduce the Few-Bit VideoQA problem, and our simple solution with feature compression. In Section~\ref{sec:experiments} we discuss several experiments to analyze and validate our approach. Finally, in Section~\ref{sec:applications} we demonstrate  applications of this novel problem, including distributing tiny versions of popular datasets and privacy.

\section{Related Work}
\label{sec:related}
\noindent \textbf{Video Question Answering}
VideoQA is a challenging task that requires the system to output answers given a video and a related question~\cite{Zhu2017UncoveringTT,jang2017tgif,xu2017video,lei2018tvqa,tapaswi2016movieqa}. Recent approaches include multi-modal transformer models~\cite{li2020hero,lei2021less} and graph convolutional networks~\cite{huang2020location}. 
We instead look at VideoQA under limited bits, which shares some philosophy with work that has looked at how much the visual content is needed for Visual Question Answering~\cite{Goyal_2017_CVPR}.

\noindent \textbf{Image and Video Coding}
Image and video coding is a widely studied problem which aims to compress image/video data with minimum loss of human perceptual quality. In the past decades, standard video codecs like HEVC \cite{sullivan2012overview}, AVC \cite{wiegand2003overview}, image codes like JPEG \cite{wallace1992jpeg} have been used for compressing image/video data. 
%which follow a hand-crafted hybrid coding framework to reduce intra- and inter-frame redundancies. 
More recently, learning-based image/video compression \cite{lu2019dvc,liu2020mlvc,hu2020imporving,abdelaziz2019neural,feng2020learned,Hu_2021_CVPR,Chadha_2021_CVPR,choi2020task,he2019beyond} has been proposed to replace the codec components with deep neural networks that optimize the entire coding framework end-to-end, to achieve better compression ratios.  
All of these existing approaches compress with the goal of pixel-level reconstruction. 
Our approach is inspired by these works but is applied under a different context --- we aim to solve a novel Few-bit VideoQA problem.

\noindent \textbf{Deep Video Representation} 
Deep neural networks have been shown effective to learn compact video representation, 
%that could be applied to various downstream tasks, 
which is now a favored way to store video data for machine learning applications. With the emergence of large-scale video datasets like Kinetics~\cite{carreira2017quo} and HowTo100M~\cite{miech19howto100m}, recent advances in representation learning \cite{feichtenhofer2019slowfast,sun2019videobert,li2020hero,miech2020end,lei2021less} extract continuous video features which contain rich semantic information. Such pre-computed video features can be successfully applied to video understanding tasks like action detection, action segmentation, video question answering, etc~\cite{li2020hero,miech2020end,lei2021less}. However, deep video features could still contain more information than what's actually needed by a specific task. 
%additionally, it's hard to interpret what information these continuous features are capturing, which might raise data privacy concerns. 
In this work, we instead focus on learning tiny video features, where we aim to use few bits of data to accomplish the target task.

\section{Few-Bit VideoQA}
\label{sec:method}

\begin{figure}[t]
\centering
\includegraphics[width=1.0\linewidth]{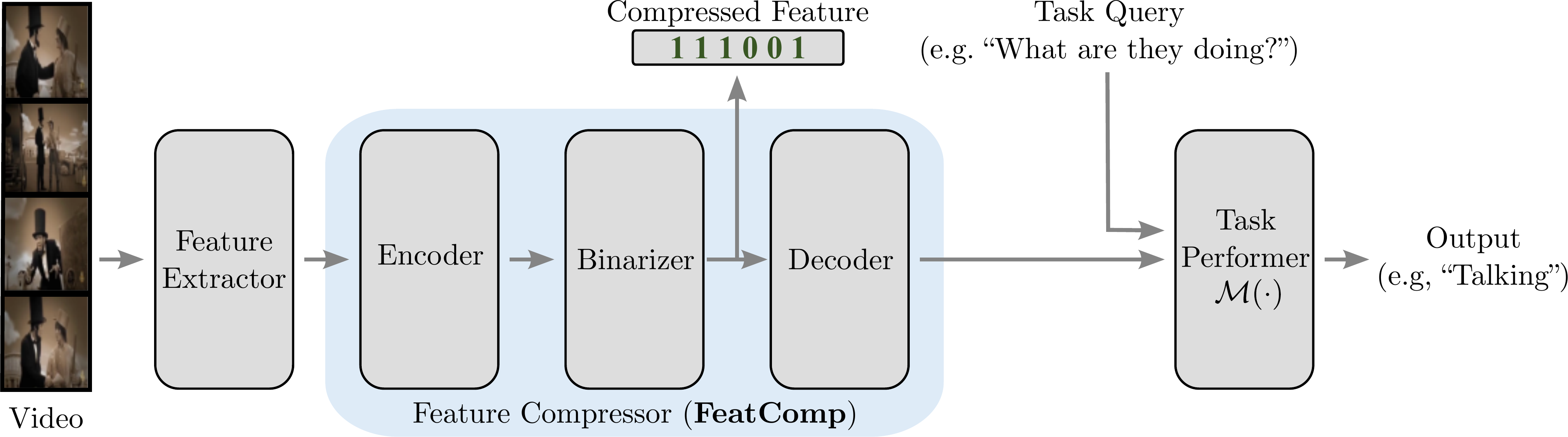}
\vspace{-2mm}
\caption{Pipeline of our generic feature compression approach (\textbf{\compressor}) towards Few-Bit VideoQA, which follows the procedure of encoding, binarization  and decoding. It has learned to only encode information relevant to the questions of interest through task-specific training.
\done{Some of the captions could be a bit more descriptive.}
}
\label{fig:pipeline_qa}
\end{figure}

In this section, we first establish the problem of Few-Bit VideoQA; then provide a simple and generic task-specific compression solution; finally we introduce our simple implementation based on a state-of-the-art VideoQA model.

\subsection{Problem Formulation}
\label{sec:prelim}

In a standard VideoQA framework, a feature extractor (e.g., ResNet3D~\cite{tran2018closer}) is applied to a video sequence to extract the video embedding $x$, which is a high-level and compact representation of the video, normally a vector composed of floating point numbers. 
%Compared to the original pixel data, a compact video feature has a smaller size and it can then be applied to perform various video tasks. 
We write $\cM(\cdot)$ as the VideoQA task performer, 
%For single modality tasks like video classification, we have the output from $\cM(x)$ as a class label; for multi-modality tasks like video question answering, 
and the output from $\cM(x, q)$ is the predicted answer in text, where $q$ refers to the text embedding of the question\done{use bold font for vectors; $\mathbf{q}$}. In our context, we assume $\cM(\cdot)$ is a neural network that can be trained with an associated task objective function $\cL_\mathrm{task}$. 

Though the compact feature $x$ already has a much smaller size compared to the original pixel data,  it still raises storage and privacy concerns as discussed in the introduction. To this end, we introduce a novel problem of Few-Bit VideoQA, where the goal is to accomplish VideoQA tasks with only few bits of visual information, i.e., we want to perform $\cM(x, q)$ with the size of $x$  less than $N$ bits, where $N$ is small (e.g., $N=10$).

\subsection{Approach: Task-Specific Feature Compression}
\label{sec:comp_module}

We propose a simple yet effective approach towards the problem of Few-Bit VideoQA. As shown in Figure~\ref{fig:pipeline_qa}, we insert a feature compression bottleneck (\textbf{\compressor}) between the video feature extractor and task performer $\cM$ to compress $x$. 
We borrow ideas from compression approaches in image/video coding~\cite{toderici2015variable,toderici2017full,wu2018video}. But rather than learning pixel-level reconstruction, we train the compression module solely from a video task loss%. And we are the first to explore compression under the context of VideoQA.
, which has not been explored before to our knowledge.
%In fact, \compressor is a generic module that can be applied to other video tasks; we will leave it for future work. 
In fact, \compressor is a generic module that could be applied to other machine learning tasks as well.

\noindent\textbf{Encoding and Decoding}
\compressor follows the encode-binarize-decode procedure to transform  floating-point features into binary, and then decode back to floating-point that can be fed into the task performer. In encoding, we first project $x$ to the target dimension, $x' = f_\mathrm{enc} (x)$,  where $x' \in \bR^N$ and $N$ is the predefined bit level to compress into.
Binarization is applied directly on feature values; so we map the feature values to a fixed range. 
We use batch normalization (BN)~\cite{ioffe2015batch} to encourage bit variance, and then a hyperbolic function $tanh(\cdot)$ to convert all the values to $[-1, 1]$.
Binarization is inherently a non-differentiable operation, in order to incorporate it in the learning process, we use stochastic binarization~\cite{toderici2015variable,toderici2017full,wu2018video} during training. The final equation for the encode-binarize-decode procedure is:
\begin{equation}
%x_{\mathrm{enc}}' = \tanh(\mathrm{BN}(x_\mathrm{enc})) 
x_\mathrm{dec} = \mathrm{\compressor}(x) = \left( f_\mathrm{dec} \ {\circ} \ \mathrm{BIN} \ {\circ} \ \tanh \ {\circ} \ \mathrm{BN} \ {\circ} \ f_\mathrm{enc} \right) (x)
\end{equation} 
where $\circ$ denotes function composition. The output after the binarization step is $x_\mathrm{bin} \in \{ 0,1\}^N$, which is the $N$-bit compressed feature to be stored.

\noindent\textbf{Learning Task-Specific Compression} Our \compressor is generic and can be inserted between any usual feature extractor and task performer $\cM(\cdot)$. The task performer will instead take the decoded feature to operate the task: $\cM(x_\mathrm{dec})$ or $\cM(x_\mathrm{dec}, q)$. To compress in a task-specific way, \compressor is trained along with $\cM(\cdot)$ with the objective $\cL = \cL_\mathrm{task}$, where $\cL_\mathrm{task}$ is the target task objective. \done{Not sure how detailed this part should be.. need any equations?} \done{I think this is fine. Let's see what our beta readers think}

\noindent\textbf{Simple Implementation}
\done{Should we use $a \times b$ instead of $a * b$?}
Our \compressor can be easily implemented into any VideoQA models. As an instantiation, we choose a recent state-of-the-art model ClipBERT~\cite{lei2021less} as our baseline, and add our compression module to study the number of bits required for VideoQA. 
Specifically, ClipBERT follows the similar pipeline as in Figure~\ref{fig:pipeline_qa}. We then insert \compressor after feature extractor. For encoding and decoding, we use a fully connected layer where $f_{\mathrm{enc}}\colon$
$\bR^{(T \times h \times w \times D)} {\mapsto} \bR^N$ and $f_{\mathrm{dec}}\colon \{0,1\}^N {\mapsto} \bR^{(T \times h \times w \times D)}$. $x$ is flattened to a single vector to be encoded and binarized, and the decoded $x_\mathrm{dec}$ can be reshaped to the original size. Then the answer is predicted by $\cM({x_\mathrm{dec}, q})$. For \compressor with $N$-bit compression, we write it as \compressor-$N$.  
This implementation is simple and generic, and as we see in Section~\ref{sec:experiments} already works surprisingly well. Various other architectures for encoding, decoding, and binarization could be explored in future research.

\noindent \textbf{Intuition of \compressor} To learn a compression of the features, various methods could be explored. We could cluster the videos or the most common answers into 1024 clusters, and encode the cluster ID in 10 bits, etc. % Our method can be interpreted in a similar fashion, but does this end-to-end, and optimizes for the clusters to be predictable from just the input.
For a task such as video action classification, where only the top class prediction is needed, directly encoding the final answer may work. However, in a general video-language task such as VideoQA, the number of possible questions is much more than 1024, even for questions about a limited topic.
Our method can be interpreted as learning $2^N$ clusters end-to-end, that are predictable, and useful to answering any questions related to the task.

\section{Experiments}
\label{sec:experiments}

In this section, we show the experimental results on Few-Bit VideoQA, study how much visual information is needed for different VideoQA tasks, and analyze what the bits capture.

\noindent \textbf{Datasets} 
We consider two public VideoQA datasets: 1) TGIF-QA~\cite{jang2017tgif}  consists 72K GIF videos, $3.0$s on average, and 165K QA pairs. We experiment on 3 TGIF-QA tasks --- Action (\eg ``What does the woman do 5 times?''), Transition (\eg ``What does the man do after talking?''), which are multiple-choice questions with $5$ candidate answers; and FrameQA (\eg What does an airplane drop which bursts into flames?''), which contains general questions with single-word answers.
2) MSRVTT-QA~\cite{xu2017video} consists of 10k  videos of duration $10{-}30$s each and 254K general QA pairs, \eg ``What are three people sitting on?''.

\noindent \textbf{Implementation Details}
We leverage the ClipBERT model pre-trained on COCO Captions~\cite{chen2015microsoft} and Visual Genome Captions~\cite{krishna2017visual}, and train on each VideoQA dataset separately, following~\cite{lei2021less}. During training, we randomly initialize \compressor, and finetune the rest of the network; we randomly sample $T{=}1$ clip for TGIF-QA and $T{=}4$ clips for MSRVTT-QA, where in each clip we only sample the middle frame. We fix the ResNet backbone and set the learning rate to $5{\times} 10^{-5}$ for \compressor, and $10^{-6}$ for $\cM$. We use the same VideoQA objective and AdamW~\cite{loshchilov2017decoupled} optimizer as in \cite{lei2021less}. During inference, we uniformly sample $T_\mathrm{test}$ clips to predict answer, where $T_\mathrm{test}{=}T$, unless noted otherwise. We set $N {=} 1,2,4,10,100,1000$ for different bit levels.
%Code and compressed datasets will be made available.
Code will be made available.

\noindent \textbf{Evaluation Metrics}
QA accuracy at different bit levels. 

\noindent \textbf{Baselines}  
\done{Baselines should probably be noindent textbf (unnumbered section)} 
To our knowledge, there is no prior work directly comparable; hence we define the following baselines:
\begin{itemize}[itemsep=2pt,topsep=0pt,parsep=0pt]
    \item \emph{Floats}: the original floating-point-based ClipBERT; it provides performance upper bound using enough bits of information~.\footnote{ClipBERT uses 16-bit precision, we use this for calculation.}
    \item \emph{Q-only}: answer prediction solely from question.
    \item \emph{Random Guess}: randomly choose a candidate/English word for multiple-choice/single-word-answer QA.
\end{itemize}
Additionally, since our approach follows an autoencoder-style design, we also study whether an objective of feature reconstruction helps with Few-Bit VideoQA. We add the following two approaches for comparison:
\begin{itemize}[itemsep=2pt,topsep=0pt,parsep=0pt]
    \item \emph{R-only}: learn \compressor solely with a reconstruction loss, $\cL_R{=} \text{MSE}(x, x_\mathrm{dec})$, then finetune $\cM$ using learned compressed features with task objective. 
    
    \item \emph{FeatComp+R}: learn \compressor from both task and reconstruction objectives, i.e., $\cL {=} \cL_{task} {+} \cL_R$.
\end{itemize}
We also study how test-time sampling affects the results with:
\begin{itemize}[itemsep=2pt,topsep=0pt,parsep=0pt]
    \item \emph{4$\times$FeatComp}: test-time sampling $T_{test}{=}4T$.

\end{itemize}

\begin{figure*}[t]
\centering
\includegraphics[width=1.0\linewidth]{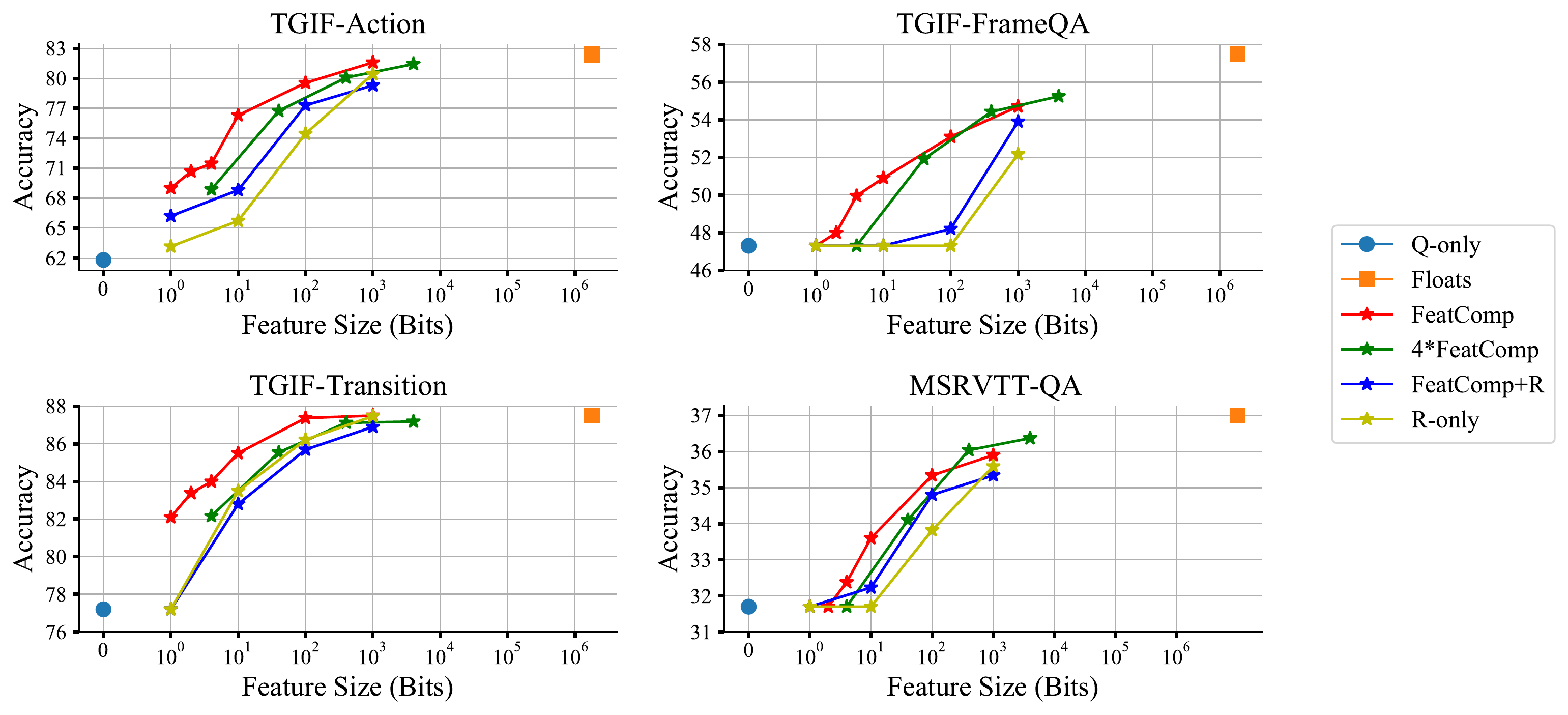}
\vspace{-4mm}

\resizebox{\textwidth}{!}{%
\begin{tabular}{cccccc}
%\begin{tabularx}{\textwidth}{Xc@{\hskip 1cm}c@{\hskip 1cm}c@{\hskip 1cm}c@{\hskip 1cm}c}
\toprule
Method & Bits & TGIF-Action & TGIF-FrameQA & TGIF-Transition & MSRVTT-QA \\
\midrule
Random Guess & 0 & 20.0 & 0.05 & 20.0 & 0.07 \\
Q-only & 0  & 61.8 & 47.3 & 77.2 & 31.7 \\
%\compressor-1 &1 & 69.0 & 46.6 & 82.1 & 30.9 \\
\taskind &  10 & 63.7 & 47.3 & 83.5 & 31.7 \\
\plusrecons &  10 & 75.8 & 47.4 & 82.8 & 32.2 \\
\compressor &10 & 76.3 & 50.9 & 85.5 & 33.6 \\
%\compressor-1000 &1000 & 81.6 & 54.7 & 87.5 & 35.9 \\
Floats &1.8M - 9.8M  & 82.4 (82.9) & 57.5 (59.4) & 87.5 (87.5) & 37.0 (37.0) \\
\midrule
 Supervision Size & &   30.9 Bits & 59.2 Bits & 57.7 Bits & 59.5 Bits \\
%\midrule
%\multicolumn{2}{c}{ Supervision Size}  & 30.9 Bits & 59.2 Bits & 57.7 Bits & 59.5 Bits \\
\bottomrule
\end{tabular}
}

\caption{Analysis of how bit size affects VideoQA accuracy on MSRVTT-QA and TGIF-QA. \emph{Floats} refers to the original ClipBERT model that serves as an upper bound of the performance without bit constraints. We both report the number we reproduced and cite paper results in parenthesis.   With our simple approach \emph{FeatComp}, we can reach high performance using only a few bits. Notably, at 10-bit level (see table), we get only $2.0{-}6.6 \%$ absolute loss in accuracy.}
\label{fig:videoqa}

\end{figure*}

\subsection{Few-Bit VideoQA Results} \label{sec:videoqa_exp}
We demonstrate our Few-Bit VideoQA results in Figure~\ref{fig:videoqa} and compare to the baselines. As expected, more bits yields accuracy improvements. Notably, for our \emph{\compressor}, even a 1-bit video encoding  yields a $7.2\%$ improvement over \emph{Q-only} on TGIF-Action. On all datasets, at 1000-bit, we maintain a performance drop ${<} 2.8 \%$ while compressing more than ${>}1000$ times. At 10-bit, the drop is within $2.0{-}6.6 \%$ while compressing over ${>}100{,}000$ times, demonstrating 10-bit visual information already provides significant aid in VideoQA tasks. 

We also compute \emph{\supervision} as a naive upper bound of bits required to accomplish the task. 
Theoretically, the compressed features should be able to differentiate the texts in all QA pairs.
Hence, for each task, we use bzip2\footnote{\url{https://www.sourceware.org/bzip2}} to compress the text file for the training QA pairs, whose size is used as the upper bound. 
We observe correspondence between \supervision and compression difficulty. TGIF-Action and TGIF-Transition are easier to compress, requiring less bits to get substantial performance gains; and they also appear to require less bits in \supervision. Instead TGIF-FrameQA and MSRVTT-QA are harder to compress, also reflected by their relatively larger \supervision.

\noindent\textbf{Role of Reconstruction Loss} 
Our approach \compressor is learned in a task-specific way in order to remove any unnecessary information. On the other hand, it is also natural to compress with the goal of recovering full data information.
Here we  investigate  whether direct feature supervision helps learn better compressed features from \emph{\plusrecons} and \emph{\taskind}. We can see that integrating feature reconstruction harms the accuracy at every bit level. In fact, \taskind can be considered as a traditional lossy data compression that tries to recover full data values. It implies that recovering feature values requires more information than performing VideoQA task; hence feature reconstruction loss may bring task-irrelevant information that hurts the task performance.

\iffalse
\begin{table}[]
\caption{Effect of adding a feature supervision $\cL_R$ (a reconstruction loss) in training objective on TGIF-QA. $\cL_R$ hurts compression ratio as it encodes task-irrelevant information. }
%\vspace{-2mm}
\centering
%\resizebox{1.0\linewidth}{!}
%{
%\renewcommand{\arraystretch}{1.0}
\begin{tabular}{ccccc}
%\begin{tabularx}{\columnwidth}{Xcccc}
%\hlineB{3}
\toprule
 & \multicolumn{2}{c}{TGIF-Action} & \multicolumn{2}{c}{TGIF-Transition} \\
\midrule
Bits & w/o $\cL_R$ & w/ $\cL_R$ & w/o $\cL_R$ & w/ $\cL_R$ \\
\midrule
1-bit set & \textbf{69.0} & 66.2  & \textbf{82.1} & 76.7 \\
10-bit set & \textbf{76.3} & 75.8 & \textbf{85.5} & 82.8 \\
1000-bit set & \textbf{81.6} & 79.3 & \textbf{87.5}  & 86.9  \\
\bottomrule
\end{tabular}
%\end{tabularx}
\label{tab:recon_loss}
\end{table}
\fi

\noindent\textbf{Role of Temporal Context}
For longer videos, frame sampling is often crucial for task accuracy. We study the impact in Figure~\ref{fig:videoqa} (\emph{\compressor} vs. \emph{4$\times$\compressor}). 
%where we have $T_\mathrm{test}{=}4$ or $16$. 
%For $T_\mathrm{test}{=}16$, we combine every $4$ evenly spaced clips to get $4$ clip bundles, compressed into $4N$ bits in total. We average the scores over the bundles for the final prediction. We additionally add the original ClipBERT result (\emph{Floats}) using $T_\mathrm{test}{=}4, 16$ clips at the end of each curve.
We can see denser sampling benefits compression ratio especially at higher bits on longer videos like MSRVTT-QA. For example, 400-bit compression with $T_\mathrm{test} {=} 16$ outperforms 1000-bit compression with $T_\mathrm{test} {=} 4$. However, on short video dataset TGIF-QA, it does not yield improvements, which implies TGIF-QA videos can be well understood with single frames.
%Sampling the same $T_\mathrm{test}{=}4$ clips,  the $1000$-bit compression model even outperforms the Floats while reducing the size by 3 orders of magnitude.

\noindent \textbf{How Task-Specific are the Features?} Here we evaluate the task-specificity by analyzing how much information is removed by the compression. We use \compressor-10 learned on a source TGIF-QA task to extract the compressed features $x_\mathrm{bin}$; then train a new VideoQA model for different target  tasks on top of $x_\mathrm{bin}$. 
For fair comparison, we use the same  $\cM$ and decoding layers and follow the previous practice to initialize $\cM$ from the model pre-trained on COCO Captions and Visual Genome, and randomly initialize the decoding layers. Then we train the network with learning rate $5{\times}10^{-5}$ for 20 epochs.
Table~\ref{tab:info_discard} shows the results, where we also cite \emph{Q-only}  from Figure~\ref{fig:videoqa}. Compression from the same task gives the best result as expected.
Note that \emph{Action} and \emph{Transition} are similar tasks in that they query for similar actions but \emph{Transition} asks change of actions over time. \emph{FrameQA} instead queries for objects, which has minimal similarity with \emph{Action}/\emph{Transition}.  
Compression from a highly relevant source task (\emph{Action} vs.\ \emph{Transition}) gives pretty high performance. However, compression from an irrelevant source task (\emph{Action}/\emph{Transition} vs.\ \emph{FrameQA}) yields similar performance as \emph{Q-only} (0-bits), which again implies that the learned compression discards any information unnecessary for its source task.

\begin{table}[t]
\caption{
Task-specificity of \compressor-10. Compression learned from an irrelevant source task is less helpful for a target task, while compression from the same task gives the best performance. 
}
%\vspace{-2mm}
\centering
%\resizebox{1.0\linewidth}{!}
% {
%\renewcommand{\arraystretch}{1.0}
\begin{tabular}{lccc}
%\begin{tabularx}{0.7\columnwidth}{Xccc}
%\begin{tabularx}{\columnwidth}{X>{\centering\arraybackslash}p{1.5cm}>{\centering\arraybackslash}p{1.5cm}>{\centering\arraybackslash}p{1.5cm}}
\toprule
& \multicolumn{3}{c}{Target Task} \\ \cmidrule{2-4}
Source Task \hspace{2em} & Action \hspace{.5em} & FrameQA \hspace{.5em} & Transition \\

%\diagbox[]{Source Task}{Target Task}  & Action & FrameQA & Transition \\
\midrule
Action & \textbf{76.9} & 47.6  & 83.8 \\
FrameQA & 62.3 & \textbf{51.7} & 77.1 \\
Transition & 76.0 & 47.3 & \textbf{85.0} \\ \midrule
Q-Only & 61.8 & 47.3 & 77.2 \\
\bottomrule
\end{tabular}
%\end{tabularx}
\label{tab:info_discard}

\end{table}

\subsection{Qualitative Analysis} 
\label{sec:qual_analysis}
\begin{figure*}[t]
\centering
     \begin{subfigure}[b]{\linewidth}
         \centering
         \includegraphics[width=\linewidth]{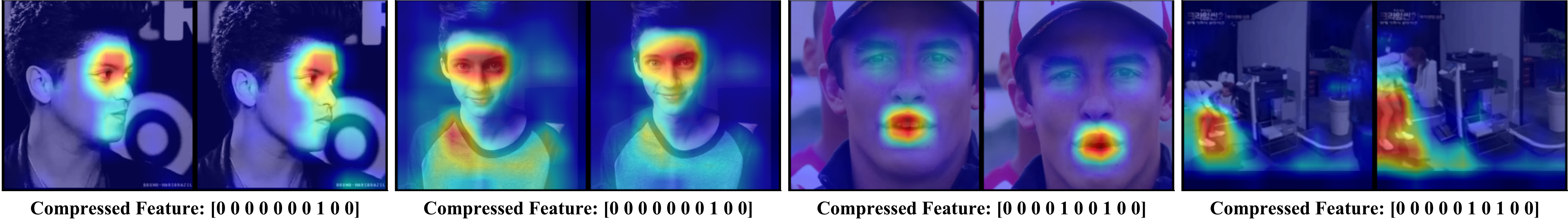}
        \caption{Bit Activation Map w.r.t. \compressor-10 compressed features $x_\mathrm{bin}$ on TGIF-Action. The learned tiny feature is able to capture salient regions useful for the source task of \compressor-10.
        }
        \label{fig:heatmap}
     \end{subfigure}
     %\hfill
     %\vspace{0.1cm}
    \begin{subfigure}[b]{\linewidth}
         \centering
         %\includegraphics[width=\linewidth]{latex/Figures/frame_cam_video0.pdf}
         %\vspace{0.3cm}
         %\includegraphics[width=\linewidth]{latex/Figures/frame_cam_video13.pdf}
        \includegraphics[width=\linewidth]{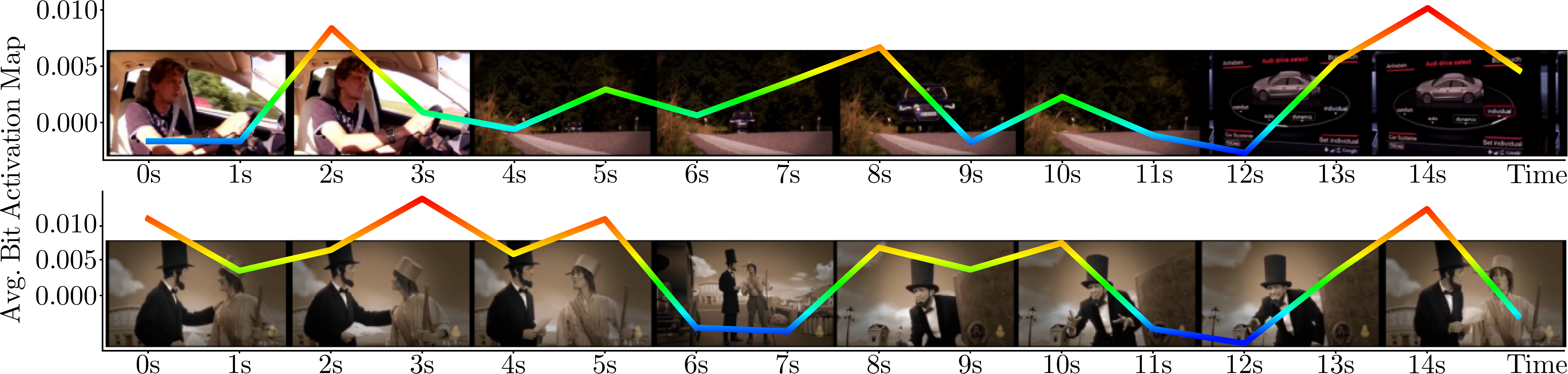}
        \caption{Averaged Bit Activation Map over video frames on MSRVTT-QA.  We can see that learned tiny feature can capture scene changes --- the bit activation peaks appear when there is a different scene useful for the source task of \compressor-10.  
        }
        \label{fig:frame_cam}
     \end{subfigure}
     %\hfill
 %\vspace{-0.8cm}
 \caption{Qualitative visualization examples of  bit activation map w.r.t \compressor-10 compressed features $x_\mathrm{bin}$.
 Note that no question, answer, prediction, or class label was used for these visualizations---This is visualizing what the network used to generically compress the video.
 \done{need to make it really clear in (a) that the questions are not used, and this is visualizing what the bits are capturing.}
 }
\label{fig:pipeline_all}
\end{figure*}

Here we study qualitatively what the learned bits are capturing. 
We apply Grad-CAM~\cite{selvaraju2017grad} directly on the compressed  features $x_\mathrm{bin}$ to find the salient regions over frames. Grad-CAM is originally a tool for localizing the regions sensitive to class prediction. It computes the weighted average of feature maps based on its gradients from the target class, which localizes the regions that contribute most to that class prediction. Here we instead construct \emph{Bit Activation Map} ({BAM}) by treating each binary bit $x^i_\mathrm{bin}$ as a ``class'': 1 is a positive class while 0 is negative. We calculate {BAM} at the last convolutional layer of ResNet. We average  {BAM} for all bits where for $x^i_\mathrm{bin}{=}0$ we multiply them with $-1$. Note that no class annotations, labels or predictions are used for {BAM}.
Figure~\ref{fig:heatmap} shows heat map visualization results on TGIF-Action for \compressor-10. The learned compression is capturing salient regions (e.g., eyes, lips, legs) which align with human perception. 
%These were determined useful for the task the compression was trained on. 
We also study in Figure~\ref{fig:frame_cam} how important temporal signals are captured, where we average the {BAM} over frames. The frames with high BAM scores tend to capture more important semantic information across time that is relevant to the task.

Additionally, we investigated whether the bits capture task-specific information by analyzing the correspondences between  VideoQA vocabularies and compressed features.
We calculate a word feature as the averaged \compressor-10 bit features over the videos whose associated QAs contain this word, and find its top-7 closest words based on  Euclidean distance. Table~\ref{tab:closest_words} shows some sample results on TGIF-Action. Neighbor words tend to demonstrate semantic associations. E.g., `eyes' is closely related to `blink', `smile' and `laugh'; `step' is associated with its similar actions like `walk' and `jump', implying that the compression captures task-relevant semantic information.

%\iffalse
%\fi

\begin{table}[]
\caption{Top-7 closest words from \compressor-10 on TGIF-Action.}
\centering
%\resizebox{1.0\linewidth}{!}
%{
%\renewcommand{\arraystretch}{1.0}
\begin{tabularx}{0.7\columnwidth}{Xl}
%\hlineB{3}
\toprule
Word & Most Similar Words \\
\midrule
head & bob neck stroke fingers raise cigarette open\\
wave & touch man bob hands hand stroke raise  \\
spin & around ice pants jump foot kick knee \\
step & walk spin around ice pants jump kick \\
guitar & object who keys black a strum bang \\
eyes & blink smile laugh cigarette lip tilt sleeve \\
\bottomrule
\end{tabularx}
%}
\label{tab:closest_words}
\end{table}

\section{Applications of Few-Bit VideoQA}
\label{sec:applications}

The problem of Few-Bit VideoQA has practical applications to data storage efficiency and privacy. Below we discuss how our approach can compress a video dataset into a tiny dataset and use that to achieve the same task (Section~\ref{sec:tiny_data}); then we demonstrate the privacy advantages of few-bit features (Section~\ref{sec:privacy}).

\subsection{Tiny Datasets} \label{sec:tiny_data}
With our task-specific compression approach, we can represent a video using only a few bits, which allows extreme compression of gigabytes of video datasets into almost nothing. For example, TGIF-QA contains 72K videos whose MPEG4-encoded format takes up about 125GB of storage; with \compressor-10, we end up with a 90KB dataset. 
We follow Section~\ref{sec:videoqa_exp} to train a model using stored compressed features $x_\mathrm{bin}$, with source and target tasks being the same, to evaluate the feature quality.
Table~\ref{tab:comp_scratch} compares the compression size and testing accuracy on TGIF-QA and MSRVTT-QA tasks. Our tiny datasets  achieve similar performance with Figure~\ref{fig:videoqa} at all bit levels. In addition to MPEG4 video and uncompressed Floats, we also report the size of Floats compressed with the off-the-shelf lossless compression standard ZIP.\footnote{\texttt{numpy.savez\_compressed}} The result implies that the original features indeed contain a lot of information not easily compressible, and we are learning meaningful compression.

\begin{table}[t]
\caption{VideoQA accuracies with our compressed datasets at different bit levels. Looking at 10-bit, we can get  100,000-fold storage efficiency, while maintaining good performance. % compared to other popular video data storage format. 
%We report the size of full datasets and report accuracy on their test splits. 
}
\centering
\resizebox{1.0\linewidth}{!}
{
\resizebox{\textwidth}{!}{
\begin{tabular}{cccccccccc}
%\begin{tabularx}{\columnwidth}{Xcccc}

%\hlineB{3}
\toprule
 & \multicolumn{4}{c}{TGIF-QA} &  & \multicolumn{4}{c}{MSRVTT-QA} \\
\cmidrule{2-5}\cmidrule{7-10}
\multirow{2}{*}{Datasets} &  \multirow{2}{*}{Size} & \multicolumn{3}{c}{Task Accs.} && \multirow{2}{*}{Size} & \multicolumn{3}{c}{Task Accs.}  \\
\cmidrule{3-5}\cmidrule{8-10}
 & & Action & FrameQA & Transition && & What & Who  & All \\
 %&
\midrule
1-bit set & 9KB & 68.3 & 47.3 & 82.4 &&  1.3KB & 24.8 & 37.7 & 31.8 \\
10-bit set & 90KB & 76.9 & 51.7& 85.0 &&  13KB & 27.4 & 44.1 & 33.7 \\
1000-bit set & 9MB & 80.8 & 54.4& 87.2 && 1.3MB  & 28.3 & 46.0 & 34.9  \\
Floats set & 16.2GB & 82.4 & 57.5 & 87.5 &&  12.3GB  &  31.7 & 45.6 &  37.0 \\
Comp. Floats set & 14.0GB  &  82.4 & 57.5 & 87.5  &&   9.5GB & 31.7 & 45.6 &  37.0 \\
MPEG4 set & 125GB &  82.4 & 57.5 & 87.5 &&   6.3GB & 31.7 & 45.6 &  37.0 \\

\bottomrule
%\hline 
%\hlineB{3}
\end{tabular}
}
%\end{tabularx}
}

\label{tab:comp_scratch}
\end{table}

\subsection{Privacy Advantages from Tiny Features}\label{sec:privacy}
Here we demonstrate how our tiny features offer privacy advantages.

\subsubsection{Advantages of Data Minimization.}

\begin{figure}[t]
\centering
\begin{minipage}{.48\textwidth}
    \centering
    %\vspace{-2mm}
    \scriptsize
    \vspace{.68cm}
    %\begin{tabular}{@{}lr@{}}
    \begin{tabularx}{1.0\linewidth}{Xr}
    \toprule
    Impossible (bits) & Possible (bits) \\ \midrule
    Credit Card Num. (${\approx}53$) & Gender (${\approx}2$) \\
    Social Security Num. (${\approx}30$) & Skin Color (${\approx}3$) \\
    Street Address (${\approx}28$) & Social Class (${\approx}3$) \\
    License Plate (${\approx}36$) & \dots \\
    Personal Image (${\approx}100,000$) & \\
    Phone Number (${\approx}33$) & \\
    \dots & \\
    \bottomrule
    \end{tabularx}%
    \captionof{table}{What sensitive information can be stored in 10 bits?}
    \label{tab:sensitive}  
\end{minipage}%
\hspace{.03\textwidth}
\begin{minipage}{.48\textwidth}
    \centering
    \includegraphics[width=1.0\linewidth]{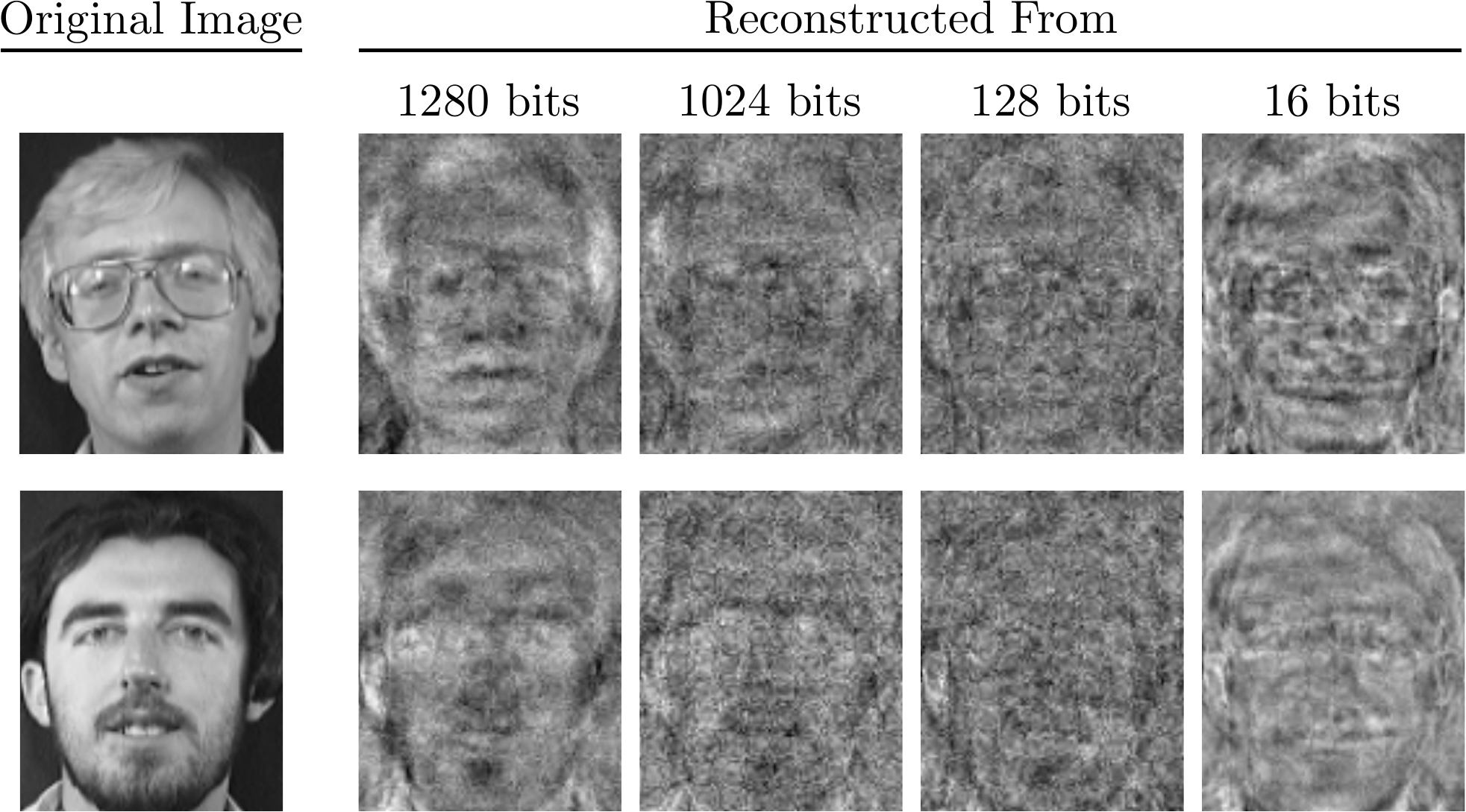}
    \vspace{-3mm}
    \captionof{figure}{Feature inversion with increasingly compressed features.}
    \label{fig:inversion}
\end{minipage}
\end{figure}

Intuitively, we expect 10 bits of information to not contain very much sensitive information. The principle of Shannon Information\footnote{\url{en.wikipedia.org/wiki/Information\_content}}, or the pigeonhole principle, tells us that 10 bits of information cannot contain a full image (approximately 10KB). Using 10 bits as an example, we can divide sensitive information into two groups based on if it can be captured by 10 bits or not in Table~\ref{tab:sensitive}.

Note we assume the data was not in the training set as otherwise the model could be used to extract that potentially sensitive information (training networks with differential privacy can relax this assumption~\cite{abadi2015deep}). By capturing only 10 bits, we can assure a user that the stored data does not contain any classes of sensitive information that require more bits to be stored.

Furthermore, considering that there are 8B unique people in the world, the identity of a person in the world can be captured in 33 bits, so we cannot reconstruct the identity of a person from 10 bits. The identity can be identifying information, photographic identity, biometrics, etc. This is consistent with the following experiment where feature-inversion techniques do not seem to work, whereas they work on regular compressed 16,384 bit features (as in Table~\ref{tbl:levels}). This effectively de-identifies the data.

\subsubsection{Robustness to Feature Inversion}
Storing only a few bits also has applications to preventing reconstruction of input from stored features (\ie Feature Inversion). Various methods exist for inverting features~\cite{mahendran2015understanding,fredrikson2015model} typically by optimizing an input to match the feature output, with an optional regularization. To evaluate this, we started with a re-implementation\footnote{\url{github.com/Koukyosyumei/secure\_ml}} of the Frederikson \etal attack~\cite{fredrikson2015model} for model inversion, but instead of class probability, we used a MSE loss between the target feature and the model feature output before the last linear layer, or after the binarizer. We used a two layer neural network trained on the AT\&T Faces Dataset~\cite{att_faces}.
In Figure~\ref{fig:inversion} we demonstrate how the inversion deteriorates as the model uses fewer bits. 
The $1{,}280$ bits result has no compression (40 32-bit numbers) whereas the rest of the results have increasing compression. All models were trained until at least $97.5\%$ accuracy on the training set.

\noindent \textbf{Feature Quantization} In Frederikson \etal~\cite{fredrikson2015model} the authors note that rounding the floating point numbers at the $0.01$ level seems to make reconstruction difficult and offer that as mitigation strategy. This suggests that compression of features will likely help defend against model inversion and we verify this here. For comparison, rounding a 32-bit float (less than $1$) at the $0.01$ level corresponds to 200 different numbers which can be encoded in $8$ bits, a 4-fold compression. Our method can compress $512$ 32-bit floats ($16{,}384$ bits) into 10 bits, a $1{,}638$-fold compression, while having a defined performance on the original task.

\subsubsection{k-Anonymity of Tiny Features}

For a system that has N users and stores 10 bits ($1{,}024$ unique values), we can assure the user that:
\begin{displayquote}
Any data that is stored is indistinguishable from the data from approximately $N/1{,}024$ other users. 
\end{displayquote}
This ensures privacy by implementing k-anonymity~\cite{Samarati98protectingprivacy} assuming the 10 bits are uniformly distributed, for example if $N{=}10^7$, $k{\approx}10{,}000$. Finally, we can use a variable number of stored bits to ensure any stored bits are sufficiently non-unique, similar to hash-based k-anonymity for password checking~\cite{li2019protocols}.

\section{Conclusion}
In this work, we introduce a novel Few-Bit VideoQA problem, which aims to do VideoQA with only few bits of video information used; we propose a simple and generic \compressor approach that can be used to learn task-specific tiny features. We experiment over VideoQA benchmarks and demonstrate a surprising finding that a video can be effectively compressed using as little as 10 bits towards accomplishing a task, providing a new perspective of understanding how much visual information helps in VideoQA. Furthermore, we demonstrate the storage, efficiency, and privacy advantages of task-specific tiny features --- tiny features ensure no sensitive information is contained while still offering good task performance. We hope these results will influence the community by offering insight into how much visual information is used by question answering systems, and opening up new applications for storing large amounts of features on-device or in the cloud, limiting privacy issues for stored features, or transmitting only privacy-robust features from a device.

\section{Acknowledgement}
We would like to thank Vicente Ordonez for his valuable feedback on the manuscript.

%We note the limitations that the study of VideoQA is relatively new, and current VideoQA datasets seem to not require much in depth understanding of the videos, and it is interesting future work to find tasks that actually require more than 10 bits to solve and how to adapt the privacy claims in this paper in those cases. The VideoQA datasets have relatively short videos, and we noticed challenges in  training models with few bits on longer videos, likely due to issues with convergence. Finally, 
%it will be important future work to understand how the
%differential-privacy concept of repeated ``draws'' (in our case, storing multiple features) will affect the privacy guarantees. \done{Another limitation is that we need a different model fro different $N$} % I wouldn't say that it's technically a limitation, more like future work.

%with applications to storing and cataloging large amounts of data for use by machine learning applications
%This approach can store large amounts of features on-device or in the cloud, limiting privacy issues for stored features, or transmitting only privacy-robust features from a device.

\clearpage
\bibliographystyle{splncs04}
\bibliography{egbib}

\begin{thebibliography}{10}
\providecommand{\url}[1]{\texttt{#1}}
\providecommand{\urlprefix}{URL }
\providecommand{\doi}[1]{https://doi.org/#1}

\bibitem{abadi2015deep}
Abadi, M., Chu, A., Goodfellow, I., McMahan, H.B., Mironov, I., Talwar, K.,
  Zhang, L.: Deep learning with differential privacy. In: Proceedings of the
  2016 ACM SIGSAC conference on computer and communications security (2016)

\bibitem{att_faces}
Cambridge, A.L.: The orl database. \url{http://www.cl.cam.ac.uk/research/dtg/
  attarchive/facedatabase.html}

\bibitem{carreira2017quo}
Carreira, J., Zisserman, A.: Quo vadis, action recognition? a new model and the
  kinetics dataset. In: CVPR (2017)

\bibitem{Chadha_2021_CVPR}
Chadha, A., Andreopoulos, Y.: Deep perceptual preprocessing for video coding.
  In: CVPR (2021)

\bibitem{chen2015microsoft}
Chen, X., Fang, H., Lin, T.Y., Vedantam, R., Gupta, S., Doll{\'a}r, P.,
  Zitnick, C.L.: Microsoft coco captions: Data collection and evaluation
  server. arXiv  (2015)

\bibitem{choi2020task}
Choi, J., Han, B.: Task-aware quantization network for jpeg image compression.
  In: ECCV (2020)

\bibitem{abdelaziz2019neural}
Djelouah, A., Campos, J., Schaub-Meyer, S., Schroers, C.: Neural inter-frame
  compression for video coding. In: ICCV (2019)

\bibitem{feichtenhofer2019slowfast}
Feichtenhofer, C., Fan, H., Malik, J., He, K.: Slowfast networks for video
  recognition. In: ICCV (2019)

\bibitem{feng2020learned}
Feng, R., Wu, Y., Guo, Z., Zhang, Z., Chen, Z.: Learned video compression with
  feature-level residuals. In: CVPR Workshops (2020)

\bibitem{fredrikson2015model}
Fredrikson, M., Jha, S., Ristenpart, T.: Model inversion attacks that exploit
  confidence information and basic countermeasures. In: Proceedings of the 22nd
  ACM SIGSAC conference on computer and communications security (2015)

\bibitem{Goyal_2017_CVPR}
Goyal, Y., Khot, T., Summers-Stay, D., Batra, D., Parikh, D.: Making the v in
  vqa matter: Elevating the role of image understanding in visual question
  answering. In: CVPR (July 2017)

\bibitem{he2019beyond}
He, T., Sun, S., Guo, Z., Chen, Z.: Beyond coding: Detection-driven image
  compression with semantically structured bit-stream. In: PCS (2019)

\bibitem{hu2020imporving}
Hu, Z., Chen, Z., Xu, D., Lu, G., Ouyang, W., Gu, S.: Improving deep video
  compression by resolution-adaptive flow coding. ECCV  (2020)

\bibitem{Hu_2021_CVPR}
Hu, Z., Lu, G., Xu, D.: Fvc: A new framework towards deep video compression in
  feature space. In: CVPR (2021)

\bibitem{huang2020location}
Huang, D., Chen, P., Zeng, R., Du, Q., Tan, M., Gan, C.: Location-aware graph
  convolutional networks for video question answering. In: AAAI (2020)

\bibitem{ioffe2015batch}
Ioffe, S., Szegedy, C.: Batch normalization: Accelerating deep network training
  by reducing internal covariate shift. In: ICML (2015)

\bibitem{jang2017tgif}
Jang, Y., Song, Y., Yu, Y., Kim, Y., Kim, G.: Tgif-qa: Toward spatio-temporal
  reasoning in visual question answering. In: CVPR (2017)

\bibitem{krishna2017visual}
Krishna, R., Zhu, Y., Groth, O., Johnson, J., Hata, K., Kravitz, J., Chen, S.,
  Kalantidis, Y., Li, L.J., Shamma, D.A., et~al.: Visual genome: Connecting
  language and vision using crowdsourced dense image annotations. IJCV  (2017)

\bibitem{Le_2020_CVPR}
Le, T.M., Le, V., Venkatesh, S., Tran, T.: Hierarchical conditional relation
  networks for video question answering. In: CVPR (2020)

\bibitem{lei2021less}
Lei, J., Li, L., Zhou, L., Gan, Z., Berg, T.L., Bansal, M., Liu, J.: Less is
  more: Clipbert for video-and-language learning via sparse sampling. In: CVPR
  (2021)

\bibitem{lei2018tvqa}
Lei, J., Yu, L., Bansal, M., Berg, T.L.: Tvqa: Localized, compositional video
  question answering. arXiv preprint arXiv:1809.01696  (2018)

\bibitem{li2020hero}
Li, L., Chen, Y.C., Cheng, Y., Gan, Z., Yu, L., Liu, J.: Hero: Hierarchical
  encoder for video+ language omni-representation pre-training. In: EMNLP
  (2020)

\bibitem{li2019protocols}
Li, L., Pal, B., Ali, J., Sullivan, N., Chatterjee, R., Ristenpart, T.:
  Protocols for checking compromised credentials. In: Proceedings of the 2019
  ACM SIGSAC Conference on Computer and Communications Security (2019)

\bibitem{liu2020mlvc}
Lin, J., Liu, D., Li, H., Wu, F.: {M-LVC}: Multiple frames prediction for
  learned video compression. In: CVPR (2020)

\bibitem{lin2019bmn}
Lin, T., Liu, X., Li, X., Ding, E., Wen, S.: Bmn: Boundary-matching network for
  temporal action proposal generation. In: ECCV (2019)

\bibitem{loshchilov2017decoupled}
Loshchilov, I., Hutter, F.: Decoupled weight decay regularization. arXiv
  preprint arXiv:1711.05101  (2017)

\bibitem{lu2019dvc}
Lu, G., Ouyang, W., Xu, D., Zhang, X., Cai, C., Gao, Z.: {DVC}: An end-to-end
  deep video compression framework. In: CVPR (2019)

\bibitem{mahendran2015understanding}
Mahendran, A., Vedaldi, A.: Understanding deep image representations by
  inverting them. In: CVPR (2015)

\bibitem{miech2020end}
Miech, A., Alayrac, J.B., Smaira, L., Laptev, I., Sivic, J., Zisserman, A.:
  End-to-end learning of visual representations from uncurated instructional
  videos. In: CVPR (2020)

\bibitem{miech19howto100m}
Miech, A., Zhukov, D., Alayrac, J.B., Tapaswi, M., Laptev, I., Sivic, J.:
  How{T}o100{M}: {L}earning a {T}ext-{V}ideo {E}mbedding by {W}atching
  {H}undred {M}illion {N}arrated {V}ideo {C}lips. In: ICCV (2019)

\bibitem{minderer2019unsupervised}
Minderer, M., Sun, C., Villegas, R., Cole, F., Murphy, K., Lee, H.:
  Unsupervised learning of object structure and dynamics from videos. arXiv
  preprint arXiv:1906.07889  (2019)

\bibitem{ILSVRC15}
Russakovsky, O., Deng, J., Su, H., Krause, J., Satheesh, S., Ma, S., Huang, Z.,
  Karpathy, A., Khosla, A., Bernstein, M., Berg, A.C., Fei-Fei, L.: {ImageNet
  Large Scale Visual Recognition Challenge}. IJCV  (2015)

\bibitem{Samarati98protectingprivacy}
Samarati, P., Sweeney, L.: Protecting privacy when disclosing information:
  k-anonymity and its enforcement through generalization and suppression
  (1998)

\bibitem{selvaraju2017grad}
Selvaraju, R.R., Cogswell, M., Das, A., Vedantam, R., Parikh, D., Batra, D.:
  Grad-cam: Visual explanations from deep networks via gradient-based
  localization. In: ICCV (2017)

\bibitem{sullivan2012overview}
Sullivan, G.J., Ohm, J.R., Han, W.J., Wiegand, T.: Overview of the high
  efficiency video coding ({HEVC}) standard. IEEE Transactions on circuits and
  systems for video technology  (2012)

\bibitem{sun2019videobert}
Sun, C., Myers, A., Vondrick, C., Murphy, K., Schmid, C.: Videobert: A joint
  model for video and language representation learning. In: ICCV (2019)

\bibitem{tapaswi2016movieqa}
Tapaswi, M., Zhu, Y., Stiefelhagen, R., Torralba, A., Urtasun, R., Fidler, S.:
  {MovieQA: Understanding Stories in Movies through Question-Answering}. In:
  CVPR (2016)

\bibitem{toderici2015variable}
Toderici, G., O'Malley, S.M., Hwang, S.J., Vincent, D., Minnen, D., Baluja, S.,
  Covell, M., Sukthankar, R.: Variable rate image compression with recurrent
  neural networks. arXiv preprint arXiv:1511.06085  (2015)

\bibitem{toderici2017full}
Toderici, G., Vincent, D., Johnston, N., Jin~Hwang, S., Minnen, D., Shor, J.,
  Covell, M.: Full resolution image compression with recurrent neural networks.
  In: CVPR (2017)

\bibitem{tran2018closer}
Tran, D., Wang, H., Torresani, L., Ray, J., LeCun, Y., Paluri, M.: A closer
  look at spatiotemporal convolutions for action recognition. In: CVPR (2018)

\bibitem{wallace1992jpeg}
Wallace, G.K.: The {JPEG} still picture compression standard. IEEE transactions
  on consumer electronics  (1992)

\bibitem{wiegand2003overview}
Wiegand, T., Sullivan, G.J., Bjontegaard, G., Luthra, A.: Overview of the
  {H.264/AVC} video coding standard. IEEE Transactions on circuits and systems
  for video technology  (2003)

\bibitem{wu2018video}
Wu, C.Y., Singhal, N., Krahenbuhl, P.: Video compression through image
  interpolation. In: ECCV (2018)

\bibitem{xu2017video}
Xu, D., Zhao, Z., Xiao, J., Wu, F., Zhang, H., He, X., Zhuang, Y.: Video
  question answering via gradually refined attention over appearance and
  motion. In: ACM MM (2017)

\bibitem{Zhu2017UncoveringTT}
Zhu, L., Xu, Z., Yang, Y., Hauptmann, A.: Uncovering the temporal context for
  video question answering. IJCV  (2017)

\end{thebibliography}
\end{document}